\documentclass{svproc}
\usepackage{url}
\usepackage[unicode=true]{hyperref}
\usepackage{graphicx}
\usepackage{subfig}

\usepackage[numbers,sectionbib]{natbib}

\mainmatter              
\makeatletter
\def\fps@figure{t}
\makeatother
\title{Agent cognition through micro-simulations: Adaptive and tunable
intelligence with NetLogo LevelSpace}
\titlerunning{Agent cognition through micro-simulation}  
\author{Bryan Head\inst{1} \and Uri Wilensky\inst{1,2}}
\authorrunning{Bryan Head and Uri Wilensky} 
\tocauthor{Bryan Head and Uri Wilensky}
\institute{
  Center for Connected Learning and Computer-Based Modeling,
  Northwestern Institute of Complex Systems,
  Department of EECS,
  Northwestern University, Evanston IL, USA\\
  Homepage: \url{https://ccl.northwestern.edu/} \\
  \email{bryan.head@u.northwestern.edu}, \email{uri@northwestern.edu}
  \and
  Department of Learning Sciences,
  Northwestern University, Evanston IL, USA
}

\begin{document}
\maketitle              

\begin{abstract}
We present a method of endowing agents in an agent-based model (ABM)
with sophisticated cognitive capabilities and a naturally tunable level
of intelligence. Often, ABMs use random behavior or greedy algorithms
for maximizing objectives (such as a predator always chasing after the
closest prey). However, random behavior is too simplistic in many
circumstances and greedy algorithms, as well as classic AI planning
techniques, can be brittle in the context of the unpredictable and
emergent situations in which agents may find themselves. Our method,
called agent-centric Monte Carlo cognition (ACMCC), centers around using
a separate agent-based model to represent the agents' cognition. This
model is then used by the agents in the primary model to predict the
outcomes of their actions, and thus guide their behavior. To that end,
we have implemented our method in the NetLogo agent-based modeling
platform, using the recently released LevelSpace extension, which we
developed to allow NetLogo models to interact with other NetLogo models.
As an illustrative example, we extend the Wolf Sheep Predation model
(included with NetLogo) by using ACMCC to guide animal behavior, and
analyze the impact on agent performance and model dynamics. We find that
ACMCC provides a reliable and understandable method of controlling agent
intelligence, and has a large impact on agent performance and model
dynamics even at low settings.
\end{abstract}

\section{Introduction}\label{introduction}

Agent-based modeling (ABM) has long proven to be a powerful method for
simulating complex systems
\citep{epstein2006generative, macal2008agent-based, wilensky2015introduction}.
Over the last decade, multi-level agent-based modeling (MLABM) has
extended this power by enabling researchers to create systems of
connected ABMs \citep{Morvan2013}. This allows one to model a system
with multiple components or levels by creating separate ABMs for each
component that are then connected. We recently released LevelSpace
\citep{Hjorth2015}, which brings the ability to integrate many different
ABMs to NetLogo \citep{wilensky1999netlogo}, one of the most widely used
ABM platforms. Here, we demonstrate how to leverage the power of MLABM
in order to define sophisticated and tunable cognitive systems for
guiding agent behavior.

Agent cognition differs from classic artificial intelligence algorithms
in that the goal is not to generate the most optimal course of action.
Instead, it is desirable to have a level of intelligence that is
appropriate for the agents in question. For instance, agents
representing humans should have significantly different cognitive
capabilities than agents representing sheep, which should, in turn, have
significantly different capabilities than agents representing ants.
Furthermore, the task is complicated by the fact that, due to the
subject matter of most ABMs, agents exist in complex environments with
many other agents, such that the patterns of their world are an emergent
result of the collective actions of those agents. Making matters more
difficult, agents are typically only aware of local information. Thus,
agents will find themselves in surprising and unexpected circumstances,
and must continue to act reasonably in those circumstances. Hence, our
goal is to design a method of agent cognition in which:

\begin{enumerate}
\def\labelenumi{\arabic{enumi}.}
\item
  Agents are given goals and information about their local state, and
  they determine what actions to take.
\item
  Agents have a tunable level of ``intelligence''.
\item
  What agents know about the world and how they think the world works is
  definable. In other words, agents should often have simplified
  understandings of the world.
\item
  Agents behave reasonably when in unexpected circumstances, or when
  something surprising occurs.
\item
  Ideally, the agents' cognitive processes should be inspectable and
  understandable. Researchers should be able to understand why agents
  are doing what they are doing.
\end{enumerate}

Our method, agent-centric Monte Carlo cognition (ACMCC), accomplishes
these goals by defining an ABM that represents how an agent thinks the
world works. By leveraging MLABM, agents then use this cognitive ABM to
make short term predictions about what will happen when they choose
different actions. The agents will then make their final action decision
based on the predicted outcomes of each action.

\section{Related Work}\label{related-work}

Under one perspective, an ABM defines a Markov chain, a small part of
which is observable by each agent. Hence, the framework of partially
observable markov decision processes (POMDPs) may be applicable in this
context \citep{lovejoy1991a}. Indeed, MDPs have been used to model
decision making of agents in dynamic environments in other works. For
instance, \citep{barve2015dynamic} combines POMDPs with reinforcement
learning to perform dynamic decision making for a single agent in a
stochastic environment. Further, \citep{littman1994markov} extends MDPs
to work with multiple agents via Markov games. MDPs are often solved by
computing the sum of discounted future reward values at each state (via
dynamic programming). However, such strategies are infeasible in the
context of ABMs as, typically, even the local area which an agent can
observe has a vast state-space, due to the number of agents and the fact
that agents are often defined by continuous attributes, and the rate at
which new information is encountered. Furthermore, even if it were
possible to find the optimal solution to the POMDPs involved, it would
not be desirable to do so, as agents typically have bounded rationality
by design. That said, our work may be characterized as using Monte Carlo
simulations to approximate the solutions to POMDPs of each agents'
surroundings.

Monte Carlo tree search (MCTS) has proven to be a particularly powerful
method to address decision processes, especially in the context of games
\citep{browne2012a}. Notably, MCTS was combined with deep reinforcement
learning techniques to create the first super-human Go playing program
\citep{silver2016mastering, silver2017mastering}. Fundamentally, MCTS
works by using Monte Carlo ``playouts'' to determine the expected value
of each possible move the player can take. It then uses those expected
values to guide future playouts, so that they focus on more promising
paths. The technique may be further augmented with other methods (such
as deep neural networks estimating the values of states) to further
guide the playouts. In this work, we take a similar Monte Carlo approach
to sampling the eventual value of actions, but use a pure Monte Carlo
strategy instead. Section~\ref{sec:futurework} discusses extending this
work to incorporate advancements in MCTS.

\citep{Head2015} also demonstrated a method of defining agent cognition
via MLABM using an early version of LevelSpace. In that work, agents
sent information about their surroundings to simple neural network
models. The output of the neural networks would then determine what
action the agents would take. This work has several advantages compared
to that, though they are complimentary methods, as discussed in
Section~\ref{sec:futurework}. First, this method requires no training.
Second, this method uses explicit cognitive representations. That is,
the objects of the agents' cognition are of the same form as the objects
of the world (entities in an ABM). When a sheep is considering the
actions of a wolf, a wolf agent exists in their cognitive model.
Furthermore, we can directly observe why an agent does what it does, and
what it expects to happen for different actions. This is in sharp
contrast with a neural network, in which the agent's knowledge is
implicitly embedded in the weights of the neurons. Finally, this method
offers a parameter that directly corresponds with the cognitive
capabilities of the agents: namely, the number of simulations the agent
runs. In contrast, the number and size of the layers in a neural network
controls the cognitive capabilities of the agents, but how that
translates into concrete differences in cognition and behavior is
opaque.

\section{Method}\label{method}

ACMCC works as follows:

\begin{enumerate}
\def\labelenumi{\arabic{enumi}.}
\item
  The modeler defines what actions an agent can take (e.g., turn left,
  turn right, go forward, eat, etc.), what the agent knows/believes
  (e.g., what the agent can see), what the agent is trying to do (e.g.,
  maximize food intake while staying alive), and how the agent thinks
  the world works via a cognitive model defined by a separate
  agent-based model. That is, this cognitive model may be a simplified
  version of the main ABM, which operates according to mechanisms as
  understood by the agent. Thus, the agent should be able to use this
  cognitive model to make predictions about what will happen in the full
  model.
\item
  During each tick of the simulation, each agent runs a settable number
  of rollouts, or short simulations of their surroundings, using its
  cognitive model, with initial conditions based on their surroundings.
  During these rollouts, the agent selects random actions and tracks how
  well they meet their objectives as a consequence of those actions. The
  agent's performance is evaluated based on reward: during a rollout,
  the agent will receive a reward based on what happens (the modeler
  defines the reward based on the agent's objectives).
\item
  The agent then selects an action based on the results of these
  rollouts.
\end{enumerate}

See Figure~\ref{fig:tree} for an example of these rollouts.

A significant advantage of this method is that it gives researchers
several tunable parameters that precisely control agents'
``intelligence'', such as the number of rollouts to run and the length
of each rollout. Having such control over agents' intelligence allows
modelers to, for instance, naturally adjust agents' cognitive
capabilities based on what is reasonable for those agents, or have an
evolvable ``intelligence'' parameter that directly correlates to the
agents' cognitive capabilities. Note that we do not claim that this is
how the actual cognitive processes of the modeled agents work.
Rather, this method simply gives modelers a practical method of
controlling the sophistication of agents' decision making abilities.

\section{Model}\label{model}

\begin{figure}
\centering
\includegraphics[width=1.00000\textwidth]{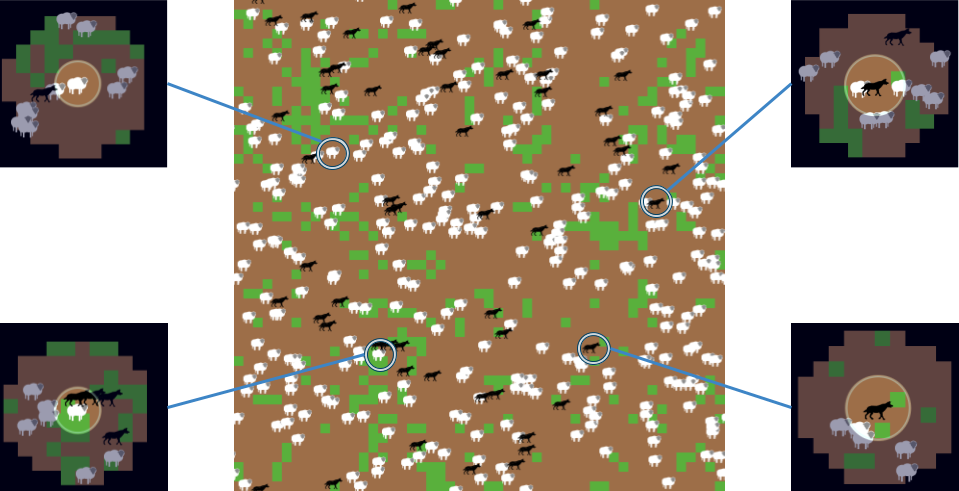}
\caption{A state from the modified Wolf Sheep Predation model. The
initial state of the cognitive model for each of the highlighted agents
is shown.}\label{fig:model}
\end{figure}

In order to demonstrate our method, we extend the Wolf Sheep Predation
model found in the NetLogo Models Library \citep{wilensky1997wsp}. Wolf
Sheep Predation is a basic predator-prey model consisting of wolves,
sheep, and grass, run on a two dimensional toroidal world. Wolves and
sheep are point-like agents that move in continuous space, while grass
exists as a binary value on a grid (either grass is present at a
particular cell or it is not). In the original model, wolves eat sheep,
sheep eat grass, and grass regrows at a fixed rate. When agents eat,
they acquire energy, and when agents move, they lose a small amount of
energy. Wolves and sheep reproduce based on a probability each tick,
splitting their energy with their child. All movement is random: each
tick, agents turn left or right by a random angle less than 45° and take
a step forward of unit length. If an agent runs out of energy, it dies
and is removed from the world. If an agent is eaten, it dies and is
removed from the world.

In our extension of the model, the random movement is replaced with
ACMCC. Agents have the following actions available to them:

\begin{enumerate}
\def\labelenumi{\arabic{enumi}.}
\item
  Turn right 30°, take a step forward of unit length, and eat if
  possible.
\item
  Take a step forward of unit length, and eat if possible.
\item
  Turn left 30°, take a step forward of unit length, and eat if
  possible.
\end{enumerate}

Reproduction still occurs based on a probability.

In order to decide what actions to take each tick, agents use a simple
cognitive model. First, agents are given a vision radius; their
cognitive model will be initialized with what they see in this radius
each tick. The cognitive model is a simplified version of the original
Wolf Sheep Predation. The world represented is much smaller than the
full model: a little bit larger than their vision radius. It is
initialized with the agents surroundings: the positions and headings of
the surroundings wolves and sheep, as well as the positions of the live
and dead grass, and finally the position, heading, and energy of the
agent itself. Furthermore, this model only includes factors of which the
agents would plausibly be aware of. The cognitive model thus makes the
following simplifications:

\begin{itemize}
\item
  Only the energy of the primary agent (called the ego) is tracked. This
  is because the ego cannot observe the energy of the other agents.
\item
  There is no reproduction. However, our model could be further extended
  to make reproduction an action, and have the child's well-being
  incorporated into the rollout reward-function. It was left out for
  simplicity and because that it is reasonable to think that the
  possibility of suddenly reproducing is not factored into the decision
  of which piece of grass to go for.
\item
  There is no grass regrowth. Again, the ego cannot observe the regrowth
  state of the grass, and these are short-term simulations.
\item
  There is nothing outside of the ego's vision radius. Should an agent
  move past the vision radius during a rollout, the grass at that point
  is filled in randomly based on the observed density of grass.
\end{itemize}

All agents in the cognitive model act randomly. While there are many
interesting possibilities for selecting actions of agents in rollouts,
such as further embedding cognitive models (thus giving agents a kind of
theory of mind), or refining rollout decision making based on past
rollouts (as in MCTS) we chose to begin with the simplest approach:
random action in rollouts. Reward in the rollouts is calculated as the
agents change in energy, with a discounting factor, so that energy
acquired in the first tick counts more than energy acquired in the
second, and so forth. Furthermore, death is given an automatic reward
value of -1000.

\begin{figure}
\centering
\includegraphics[width=0.81000\textwidth]{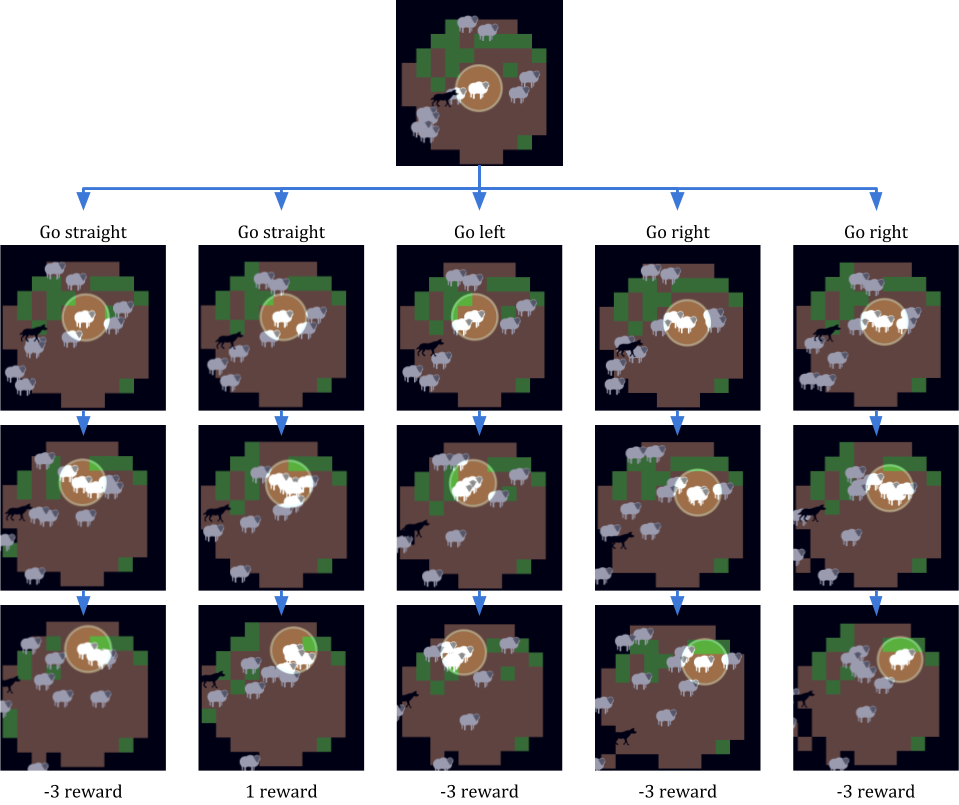}
\caption{The tree of rollouts performed by the highlighted sheep, which
is facing north. The rollouts progress from top to bottom, with the
first row showing the state after 1 tick, and the last row showing the
final state. The sheep's initial actions and final reward values are
shown (without discounting).}\label{fig:tree}
\end{figure}

The main predator-prey model and cognitive model are then combined as
follows. Each tick of the predator-prey model proceeds as in the
original Wolf Sheep Predation, save that agents send information about
their surroundings to their cognitive model, which then runs a set
number of rollouts of a set length, and reports the resulting reward
values. The agent then takes the action that had the highest mean
reward. See Figure~\ref{fig:model} for a state of the predator-prey
model and the corresponding initial state of several of the cognitive
models. Figure~\ref{fig:tree} shows the various rollouts performed by
the cognitive model of a sheep at that tick. Through these rollouts, the
sheep finds that the two sheep coming from the east almost always eat
the patch of grass to the northeast, and the sheep coming from the north
almost always eat the grass to the north-northwest. Thus, the only grass
the sheep manages to successfully eat is the northernmost (final image
in the second column). Thus, the sheep decides to go straight. In this
way, ACMCC allows for agents to perform sophisticated decision making
where a greedy algorithm would have failed (because the closest grass
would have been eaten by other sheep first).

\section{Results}\label{results}

In order to understand the effects of the cognitive model on Wolf Sheep
Predation, several experiments were performed. In each experiment, one
agent-type had the number and length of its rollouts varied while the
other agent-type was kept random. The efficiency of the agents in
various tasks as well as population levels were recorded. Two
measurements of efficiency were used: one for sheep and one for wolves.
Sheep efficiency was measured as the ratio of the amount of grass eaten
in a tick to the amount of grass we would expect to be eaten by random
sheep, given grass density and number of sheep:
\texttt{grass-eaten\ /\ (num-sheep\ *\ grass-density)}. Similarly, wolf
efficiency was measured as the ratio of the amount of sheep eaten in a
tick to the amount of sheep we would expect to be eaten by random
wolves, given sheep density and number of wolves:
\texttt{sheep-eaten\ /\ (num-wolves\ *\ sheep-density)}. This measure
was used as it controls for population dynamics: it measures how well
the agents are doing compared to how well we would expect random agents
to be doing. Thus, in both cases, an efficiency of 1 is random. However,
as there are relatively few wolves (in the base model, wolves oscillate
between 50 and 100, while sheep oscillate between 120 and 200) with
relatively few prey (the sheep's prey, grass, covers around 1000
locations at any given tick), this number rests below 1 for the baseline
case (at around 0.76) due to the fact that wolves and their prey are
discrete entities. Sheep, on the other hand, are more numerous with an
evenly spread food source, and thus their baseline rests close to the
theoretical value of 1, at 0.93.

\begin{figure}
\centering

\subfloat[Sheep using rollouts, wolves
random]{\includegraphics[width=0.470000\textwidth]{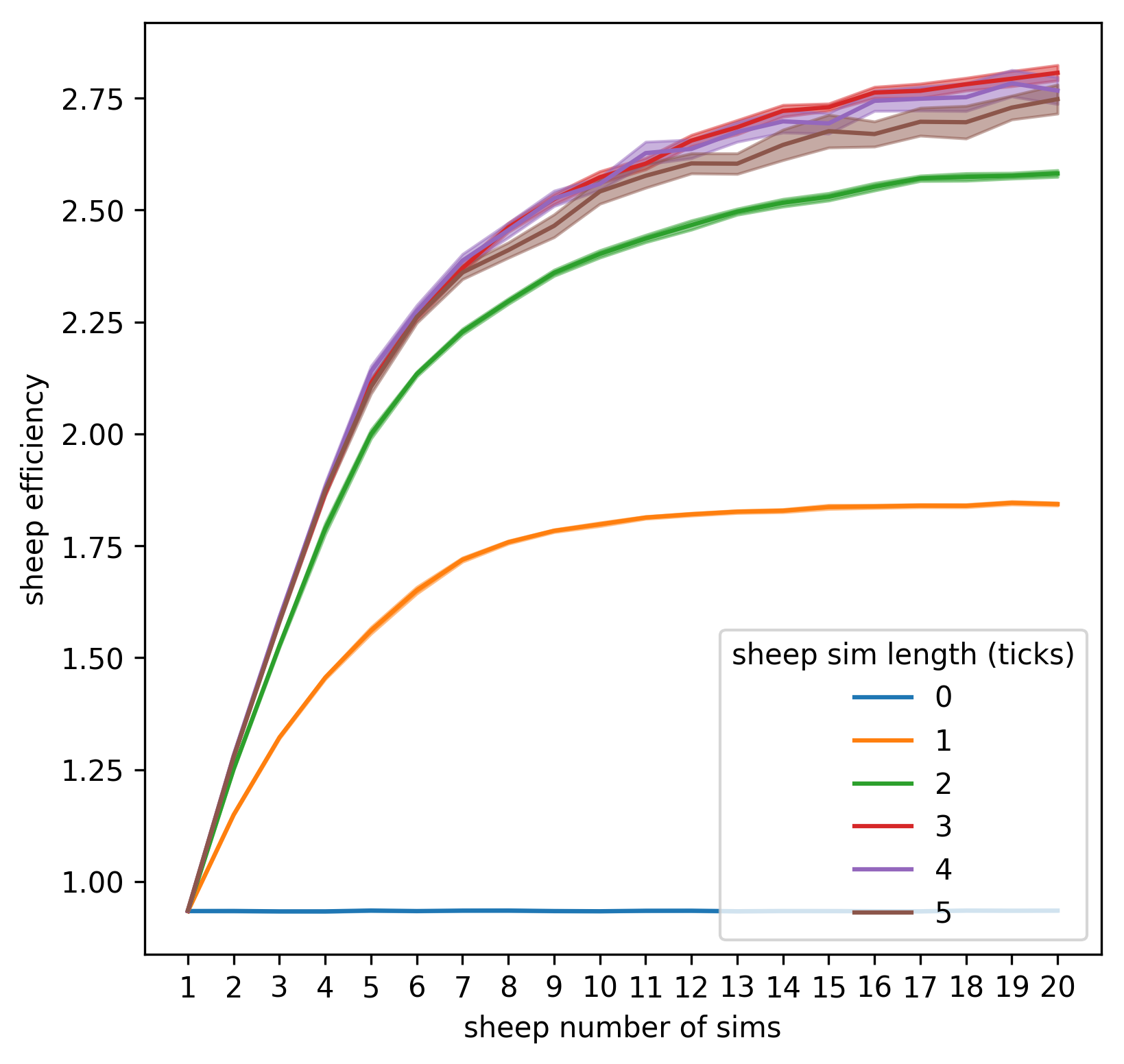}}\quad
\subfloat[Wolves using rollouts, sheep
random]{\includegraphics[width=0.470000\textwidth]{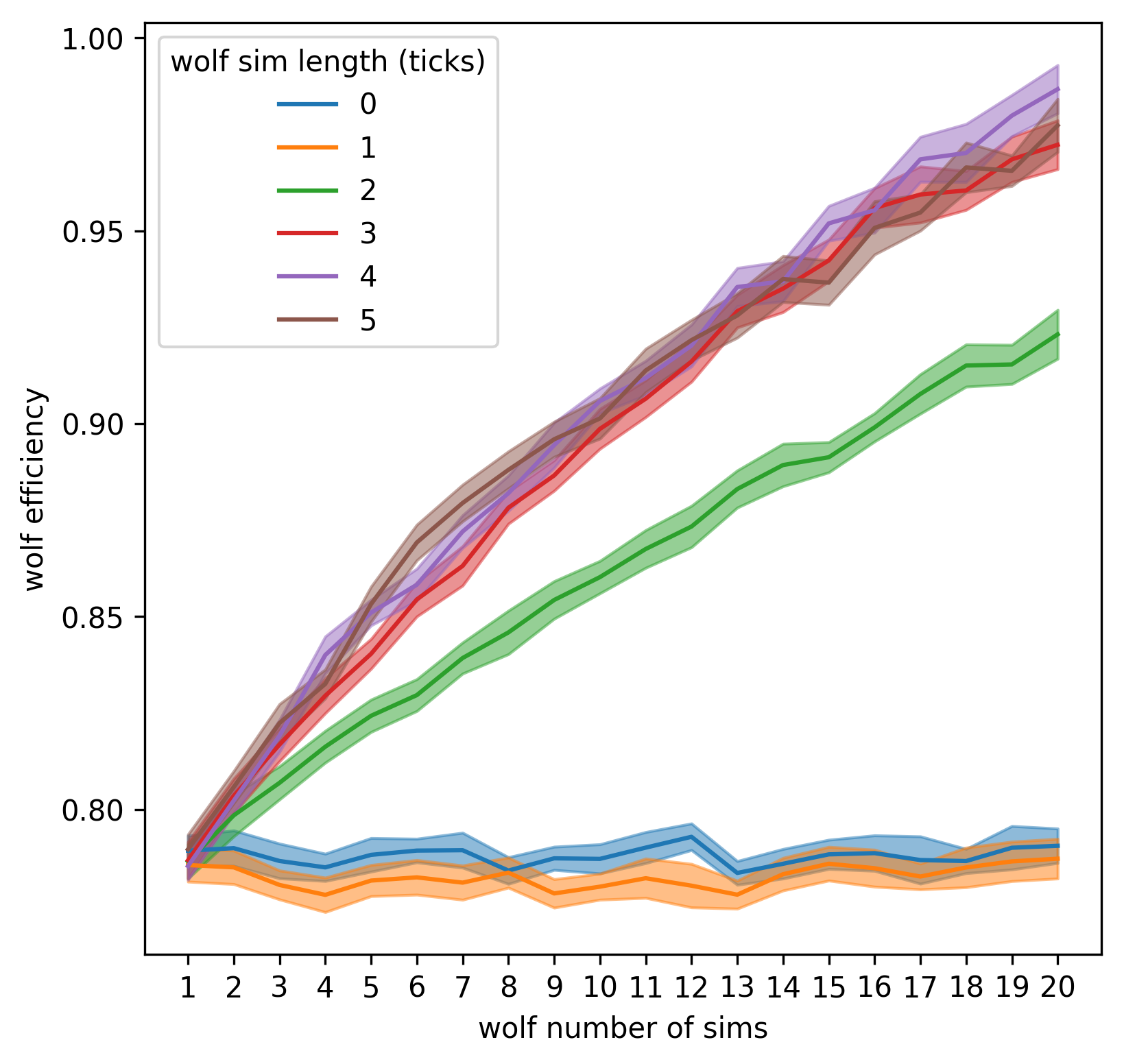}}

\caption{Mean efficiency of sheep and wolves at eating their respective
foods (grass and sheep) for various combinations of number and length of
rollouts. Bands show 95\% confidence intervals. Note that the y-axes
have different scales.}

\label{fig:efficiency}

\end{figure}

For both wolves and sheep, number of rollouts was varied from 1 to 20
(note that performing only a single rollout is equivalent to random
behavior) and the length of rollouts was varied from 0 to 5. Vision
radius was set to 5 for both agent types. Default parameter settings (as
found in the NetLogo Models Library) were used for all other parameters.
Runs were carried out for 2,000 ticks. 20 repetitions were performed for
each combination of parameters. Figure~\ref{fig:efficiency} shows the
efficiency results for both agent-types.

\begin{figure}
\centering

\subfloat[Sheep using rollouts, wolves
random]{\includegraphics[width=0.47000\textwidth]{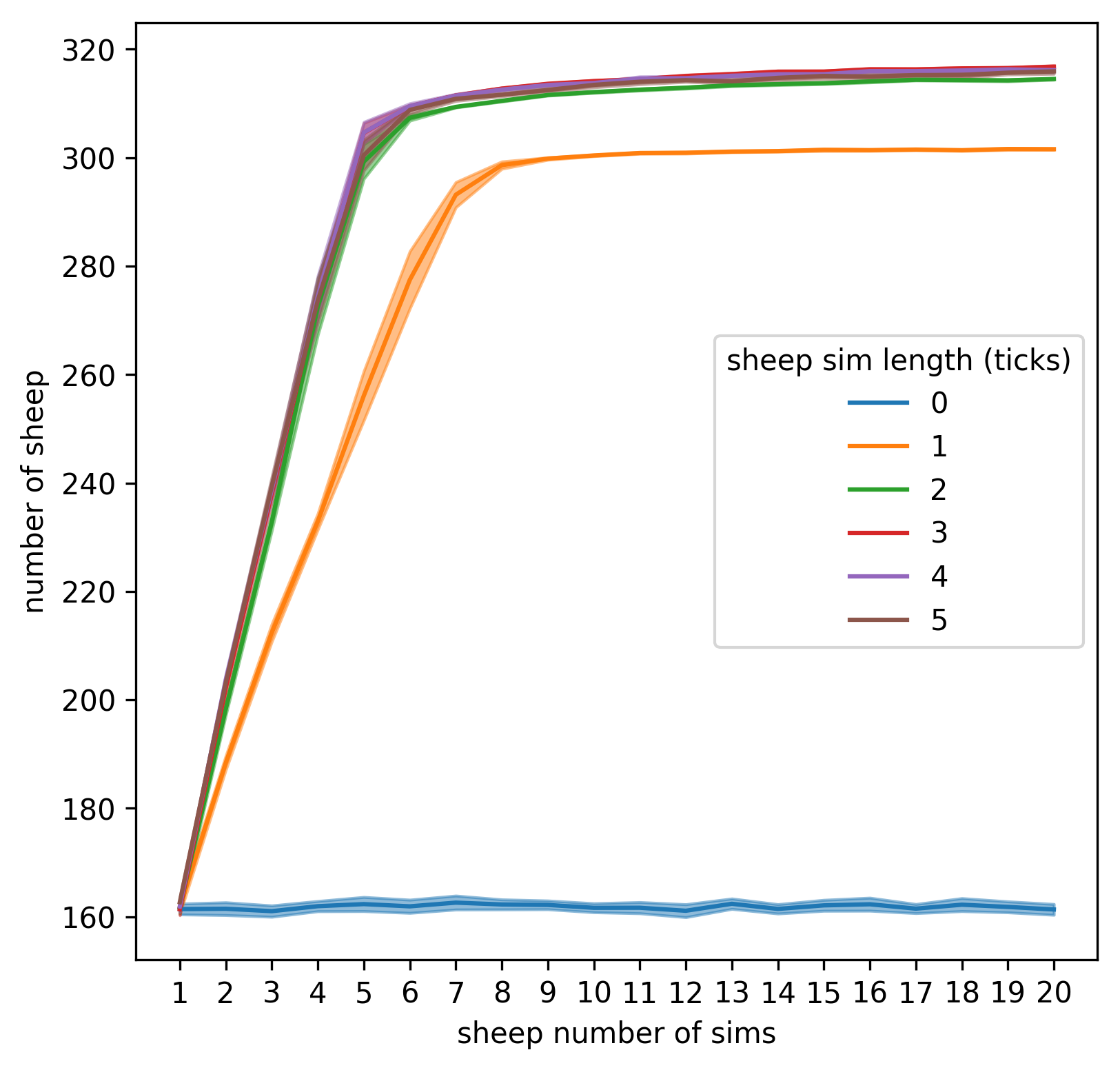}\label{fig:sheeppop-vs-sheep}}\quad
\subfloat[Wolves using rollouts, sheep
random]{\includegraphics[width=0.47000\textwidth]{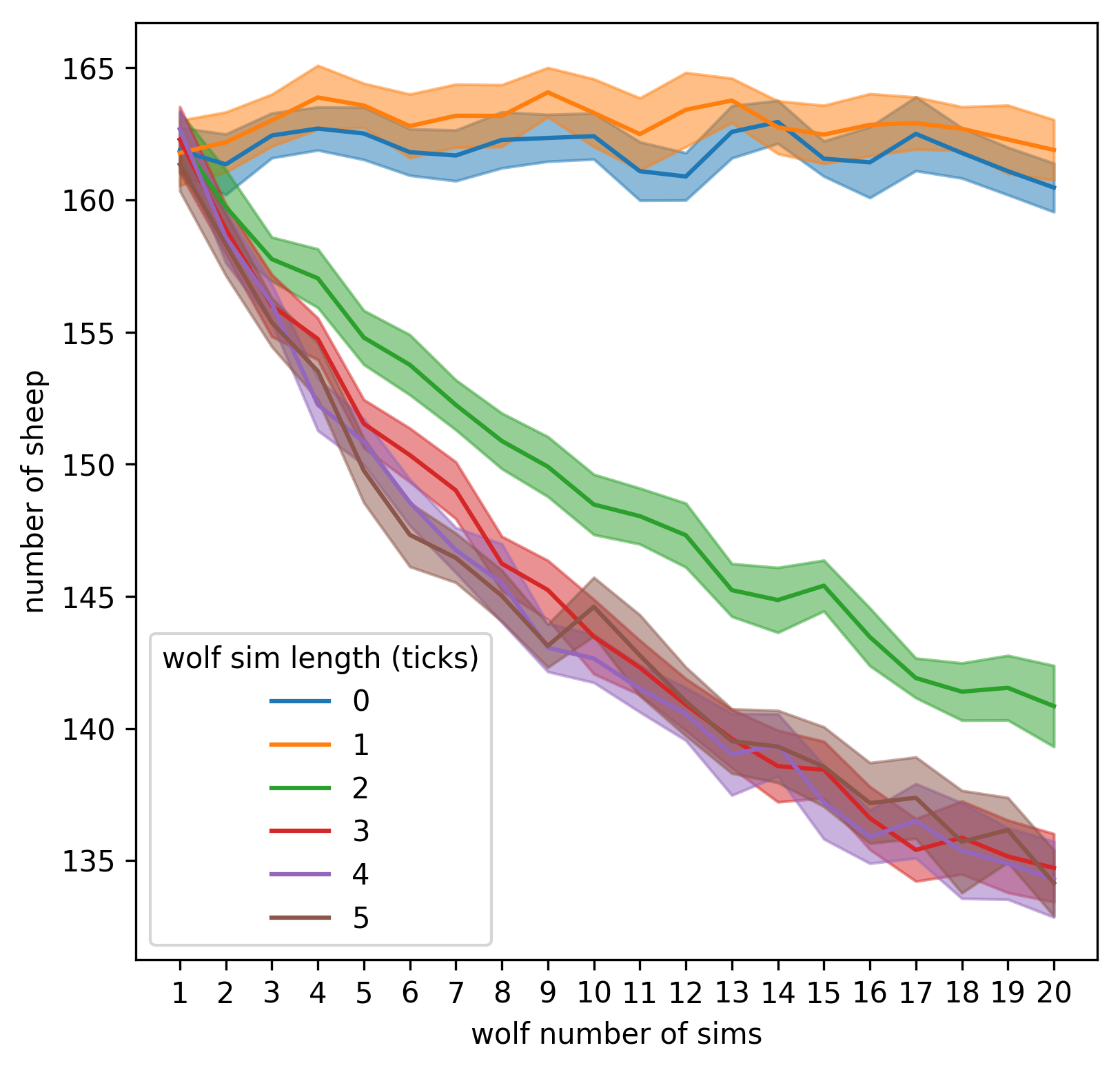}\label{fig:sheeppop-vs-wolf}}

\caption{Mean sheep population for various combinations of number and
length of rollouts performed by sheep and wolves. Bands show 95\%
confidence intervals. Note that the y-axes have different scales.}

\label{fig:sheeppop}

\end{figure}

Finally, in order to begin to understand the impact of the cognitive
model on the dynamics of the model, we examine the sheep population.
Using the same runs as above, Figure~\ref{fig:sheeppop-vs-sheep} shows
the mean population of sheep for each combination of number and length
of sheep rollouts. In contrast, Figure~\ref{fig:sheeppop-vs-wolf} shows
mean sheep population in response to wolf cognitive abilities. Means are
taken from tick 500 onwards in each run, to allow for the system to
reach stability.

\section{Discussion}\label{discussion}

We find that the number of simulations has a direct and monotonic impact
on agent performance, and thus works well as an ``intelligence''
parameter. Furthermore, even a small number of short rollouts
dramatically increases performance in sheep. Thus, it is possible to
give hundreds of agents fairly sophisticated short-term reasoning in
this way without too high of a performance impact: all simulations
performed here are runnable in real-time on a standard, modern laptop.
Rollout length appeared to reach maximum efficacy at around three ticks.
This is not surprising; the longer rollouts are run for, the less
accurate they will be, especially as agents could only see to distance
5. This invites the question of how to do longer term predictions about
the future; this is discussed in Section~\ref{sec:futurework}.

While wolves were not helped as much as sheep, their performance was
still improved with a relatively small number of short rollouts.
Regardless, their poorer performance is not surprising: catching a
moving prey is much more difficult than catching a stationary one, or
even than avoiding being caught. Nevertheless, wolf performance is
impressive considering the branching factor and their poor model of
sheep behavior. Methods for overcoming their natural disadvantage are
discussed in Section~\ref{sec:futurework}.

As shown in Figure~\ref{fig:sheeppop-vs-wolf}, as the wolves become
smarter, their food supply significantly drops. This is particularly
significant as fewer sheep means fewer sheep reproducing, and thus less
food for the wolves being introduced each tick. This is unlike the
sheep's food source, which regrows at each location at a fixed rate, and
thus, the fewer locations occupied by grass, the faster it is introduced
into the system. The effect of wolf behavior on sheep population
highlights the difficulties of applying cognitive systems; due to the
aggregate-level feedback loop in the system, what's beneficial to the
individual can be harmful to the group.

More broadly, these results indicate that this is a promising strategy
for giving agents more sophisticated, yet tunable cognitive capabilities
in a natural way, using nothing but agent-based modeling.

\section{Future work}\label{sec:futurework}

While this work lays a solid foundation for a novel approach to agent
cognition, it can be extended in many interesting ways.

First, as the agents are only performing a handful of rollouts, it is
important that those rollouts focus on promising/likely actions and
futures. This is particularly salient in the case of the wolves, who
have the more difficult task of catching a moving target. There are a
number of established ways of accomplishing this. First, in MCTS, this
is accomplished by using the results of past rollouts to guide future
rollouts. There are some difficulties here: with the few number of
rollouts and the continuous state-space of many ABMs, it is unlikely for
the rollouts to encounter the same state twice. Regardless, this method
is immediately useful for the choice of the first action, and could be
modified to work on future actions by either compressing states (using a
learned encoder) or scoring sequences of actions, regardless of state.

Another method that has recently proven to be highly effective is
combining MCTS with neural networks, as was done in AlphaGo
\citep{silver2016mastering} and AlphaGo Zero
\citep{silver2017mastering}. In ABM, neural networks could be trained to
both select likely actions for the ego, as well as to predict the
behavior of the other agents. This method would both improve the
efficacy of rollouts, and offer a way of incorporating learning into the
system, if that is desirable. Our initial work in this direction has
been promising, which combines this work with the work in
\citep{Head2015}.

Another method of better predicting the actions of other agents would be
to embed another layer of ACMCC models inside the first layer. That is,
agents would have a kind of theory of mind, where they try to emulate
the thinking of the other agents based on what they can observe. The
drawback to the naive implementation of this, however, is that it is
performance intensive (initial experiments reinforce this). However,
this may be circumvented by doing a kind of MCTS for each agent at the
high level; that is, each agent in the cognitive model uses the results
of past rollouts to improve their behavior in future rollouts, thus
emulating the cognition of the other agents without performing any
additional rollouts. Regardless, the naive approach of further embedding
cognitive models could be quite effective for models with fewer agents
that require highly sophisticated reasoning. For instance,
\citep{rabinowitz2018machine} examines an approach to simulating theory
of mind in a multi-agent simulation somewhat along these lines.

Another challenge is adapting this method to work with continuous action
spaces. As described here, the actions must be discrete; wolves and
sheep can only turn by fixed amounts, while they can turn by a
continuous amount in the original model. This could be accomplished by
interpolating between different discrete parameters for an action.

While the method as applied here appears to be effective for short-term
reasoning, it does not perform any kind of long term reasoning. A simple
way of adapting the method to perform longer term reasoning would be to
decrease the accuracy of the simulation while increasing it's speed by
changing its timestep. Another method would be to have the cognitive
model operate on a coarser grain than the main model.

Thus, while this work lays the foundation for sophisticated agent
cognition, it opens up many possible avenues of exploration as well.

\bibliography{bib}

%
\end{document}